\documentclass{Interspeech}
\usepackage{algorithm}
\usepackage{algpseudocode}
\usepackage{listings}
\usepackage{minted}
\usepackage{caption}
\usepackage{graphicx}
\usepackage[T1]{fontenc}
\usepackage[utf8]{inputenc}
\usepackage{float}
\usepackage{amssymb}
\usepackage{pifont}
\newcommand{\cmark}{\ding{51}}
\newcommand{\xmark}{\ding{55}}

\interspeechcameraready

\title{The ML-SUPERB 2.0 Challenge: \\
Towards Inclusive ASR Benchmarking for All Language Varieties}

\author[affiliation={1}]{William}{Chen}
\author[affiliation={2}]{Chutong}{Meng}
\author[affiliation={1}]{Jiatong}{Shi}
\author[affiliation={3}]{Martijn}{Bartelds}
\author[affiliation={1}]{Shih-Heng}{Wang}
\author[affiliation={4}]{Hsiu-Hsuan}{Wang}
\author[affiliation={5,8}]{Rafael}{Mosquera}
\author[affiliation={5,8}]{Sara}{Hincapie}
\author[affiliation={3}]{Dan}{Jurafsky}
\author[affiliation={2,7}]{Antonis}{Anastasopoulos}
\author[affiliation={4}]{Hung-yi}{Lee}
\author[affiliation={6}]{Karen}{Livescu}
\author[affiliation={1}]{Shinji}{Watanabe}

\affiliation{}{Carnegie Mellon University, $^2$George Mason University, $^3$Stanford University, $^4$National Taiwan
                 University, $^5$ML Commons, $^6$TTI-Chicago, $^7$Archimedes, Athena RC, $^8$Factored AI}{} 
\email{williamchen@cmu.edu}

\usepackage[
backend=biber,
style=ieee,
citestyle=numeric-comp,
maxbibnames=3,
maxcitenames=3,
doi=false,isbn=false,url=false,eprint=false
]{biblatex}
\defbibheading{bibliography}[\refname]{}

\DeclareSourcemap{
	\maps[datatype=bibtex, overwrite=true]{
		\map{
			\step[fieldsource=booktitle,
			match=\regexp{.*Interspeech.*},
			replace={Interspeech}]
			\step[fieldsource=journal,
			match=\regexp{.*INTERSPEECH.*},
			replace={Interspeech}]
			\step[fieldsource=booktitle,
			match=\regexp{.*ICASSP.*},
			replace={ ICASSP}]
			\step[fieldsource=booktitle,
			match=\regexp{.*icassp_inpress.*},
			replace={ ICASSP (in press)}]
			\step[fieldsource=booktitle,
			match=\regexp{.*Acoustics,.*Speech.*and.*Signal.*Processing.*},
			replace={ ICASSP}]
			\step[fieldsource=booktitle,
			match=\regexp{.*International.*Conference.*on.*Learning.*Representations.*},
			replace={ ICLR}]
			\step[fieldsource=booktitle,
			match=\regexp{.*International.*Conference.*on.*Computational.*Linguistics.*},
			replace={ COLING}]
			\step[fieldsource=booktitle,
			match=\regexp{.*SIGdial.*Meeting.*on.*Discourse.*and.*Dialogue.*},
			replace={ SIGDIAL}]
			\step[fieldsource=booktitle,
			match=\regexp{.*International.*Conference.*on.*Machine.*Learning.*},
			replace={ ICML}]
			\step[fieldsource=booktitle,
			match=\regexp{.*North.*American.*Chapter.*of.*the.*Association.*for.*Computational.*Linguistics:.*Human.*Language.*Technologies.*},
			replace={ NAACL}]
			\step[fieldsource=booktitle,
			match=\regexp{.*Empirical.*Methods.*in.*Natural.*Language.*Processing.*},
			replace={ EMNLP}]
			\step[fieldsource=booktitle,
			match=\regexp{.*Association.*for.*Computational.*Linguistics.*},
			replace={ ACL}]
			\step[fieldsource=booktitle,
			match=\regexp{.*Automatic.*Speech.*Recognition.*and.*Understanding.*},
			replace={ ASRU}]
			\step[fieldsource=booktitle,
			match=\regexp{.*Spoken.*Language.*Technology.*},
			replace={ SLT}]
			\step[fieldsource=booktitle,
			match=\regexp{.*Speech.*Synthesis.*Workshop.*},
			replace={ SSW}]
			\step[fieldsource=booktitle,
			match=\regexp{.*workshop.*on.*speech.*synthesis.*},
			replace={ SSW}]
			\step[fieldsource=booktitle,
			match=\regexp{.*Advances.*in.*neural.*information.*processing.*},
			replace={ NeurIPS}]
			\step[fieldsource=booktitle,
			match=\regexp{.*Advances.*in.*Neural.*Information.*Processing.*},
			replace={ NeurIPS}]
			\step[fieldsource=booktitle,
			match=\regexp{.*Workshop.*on.* Applications.* of.* Signal.*Processing.*to.*Audio.*and.*Acoustics.*},
			replace={ WASPAA}]
            \step[fieldsource=booktitle,
			match=\regexp{.*Language.*Resources.*and.*Evaluation.*Conference.*},
			replace={ LREC}]
			\step[fieldsource=publisher,
			match=\regexp{.+},
			replace={{}}]
			\step[fieldsource=month,
			match=\regexp{.+},
			replace={{}}]
			\step[fieldsource=location,
			match=\regexp{.+},
			replace={{}}]
			\step[fieldsource=address,
			match=\regexp{.+},
			replace={{}}]
			\step[fieldsource=organization,
			match=\regexp{.+},
			replace={{}}]
		}
	}
}
\keywords{multilingual, speech recognition}

\usepackage{comment}

\begin{document}

\maketitle

\begin{abstract}
    Recent improvements in multilingual ASR have not been equally distributed across languages and language varieties. To advance state-of-the-art (SOTA) ASR models, we present the Interspeech 2025 ML-SUPERB 2.0 Challenge. We construct a new test suite that consists of data from 200+ languages, accents, and dialects to evaluate SOTA multilingual speech models. The challenge also introduces an online evaluation server based on DynaBench, allowing for flexibility in model design and architecture for participants. The challenge received 5 submissions from 3 teams, all of which outperformed our baselines. The best-performing submission achieved an absolute improvement in LID accuracy of 23\% and a reduction in CER of 18\% when compared to the best baseline on a general multilingual test set. On accented and dialectal data, the best submission obtained 30.2\% lower CER and 15.7\% higher LID accuracy, showing the importance of community challenges in making speech technologies more inclusive.
\end{abstract}

\section{Introduction}

In the past decade, studies on scaling end-to-end neural networks have led to dramatic improvements in models for Automatic Speech Recognition (ASR) \cite{baevski2020wav2vec, hsu2021hubert}. Importantly, ASR systems are no longer limited to solely the English language: state-of-the-art (SOTA) models achieve strong performance on over 50 languages \cite{whisper, owsm, usm}.

However, these benefits have not been equally distributed among all languages and language varieties. Many works have shown that SOTA models perform significantly worse on accented speech or dialects that are not considered ``standard" \cite{changAAVE, feng2024towards, iwslt}. This causes many downstream technologies, such as virtual assistants, closed captioning, and translation services to function substantially worse for many subgroups that may already be disadvantaged.

To address this limitation, this paper presents the Interspeech 2025 ML-SUPERB 2.0 Challenge. The goal of the challenge is to encourage the development of systems that can perform well across languages and language varieties. In total, the challenge evaluates ASR models on speech data from \textit{149 languages and 93 language varieties}. Unlike other SUPERB-based benchmarks and challenges \cite{superbsg, avsuperb, mlsuperb, feng2023superb, mlsuperbchallenge, shi2024mlsuperb20benchmarkingmultilingual, yang21c_interspeech}, which focused on smaller probe-based experiments on limited amounts of data, the ML-SUPERB 2.0 Challenge has no restrictions on what data or pre-trained models can be used. Participants can curate training data or leverage the latest advances in speech modeling to train the best possible models. Evaluation of these models is facilitated by our online evaluation server that is hosted by DynaBench \cite{dynabench}. This server performs inference for the participants by having all submissions adhere to an API. As such, the test set remains fully hidden to participants, preventing benchmark overfitting.

The focus on robustness to \textit{both} different languages and language varieties is novel, as research in these areas has often been disjoint. Many works aim to solely increase the amount of languages a system can handle, but ignore variation that occurs within languages \cite{mms, usm, chen2024towards}. On the other hand, research on improving ASR performance on different accents or dialects often focuses on variation within single languages \cite{changAAVE, artie, accentarch, fairspeech}. This benchmark acts as a first step towards better unifying the fields of \textit{multilingual} and \textit{fair} speech processing.  We hope such cross-disciplinary interactions can lead to both a more inclusive speech processing community and more inclusive speech processing models. This paper is outlined as follows:

\begin{itemize}[left=6pt,nolistsep,noitemsep]
    \item Section \ref{sec:data} details the collection and cleaning process for the data used in the challenge, while discussing the difficulties in developing a benchmark with such a wide coverage of languages and language varieties.
    \item Section \ref{sec:benchmark} outlines the challenge rules and design decisions.
    \item  Section \ref{sec:experiments} presents baseline results with SOTA self-supervised and supervised ASR models, and compares them to the submitted systems. 
\end{itemize}
Our contributions can thus be summarized as:
\begin{enumerate}[left=6pt,noitemsep,nolistsep]
    \item We introduce a new challenge that evaluates multilingual ASR performance across 149 languages and 93 language varieties, representing the broadest coverage of any speech benchmark to date.
    \item We compare 5 submitted systems, which all out-performed our baseline systems, showing that community challenges can lead to better-performing systems.
    \item Despite these advancements, we find that SOTA ASR systems continue to underperform on accented and dialectal speech.
\end{enumerate}
\begin{figure*}[ht]
    \centering
    \includegraphics[width=1\linewidth]{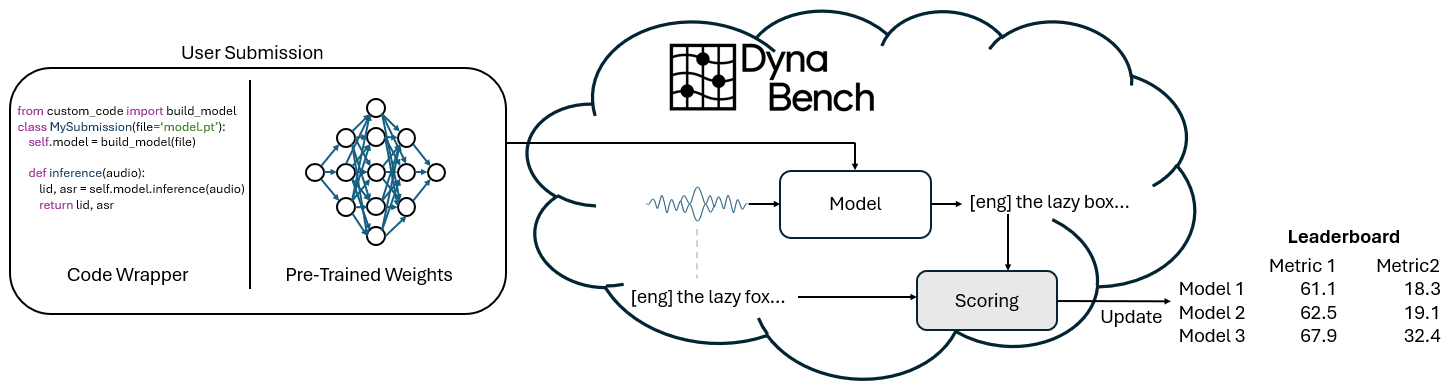}
    \caption{\textbf{Overview of the challenge submission system.} Participants upload their model weights and inference code to DynaBench, which runs the evaluation online and returns the evaluation metrics.}
    \vspace{-0.5cm}
    \label{fig:dynabench}
\end{figure*}

\section{Challenge Data} \label{sec:data}

\subsection{General Multilingual Data} \label{sec:multilingual_data}
The data described in this section is designed to evaluate the general multilingual capabilities of ASR models across 149 languages. We obtain this data by combining previous ML-SUPERB benchmarks \cite{mlsuperb, mlsuperbchallenge,  shi2024mlsuperb20benchmarkingmultilingual}. In doing so, we found several issues with the data used in these existing benchmarks or the corpora they were sourced from. Many of these issues originate from cases where a language may use several different orthographies. For example, we removed all Norwegian data (\texttt{nor}, \texttt{nno}, and \texttt{nob}), because (1) some data sources do not clearly indicate which written standard they represent; (2) \texttt{nno} and \texttt{nob} differ only in writing systems and they do not have a clear relationship with the spoken dialects. Similar issues were observed in Serbian: FLEURS \cite{conneau2023fleurs} had Latin transcripts while Common Voice \cite{ardila2020common} used Cyrillic. Normalizing these factors is particularly important for our challenge, as otherwise our metrics on fairness and robustness will be severely impacted (Section \ref{sec:metrics}). Another common problem was the distinction between similar languages. Although Tagalog (\texttt{tgl}) and Filipino (\texttt{fil}) are technically distinct languages, the source datasets do not explicitly distinguish them. Therefore, we decided to merge them. Also, the choice of ISO code used by different datasets posed a problem, where some datasets use the macro-language designation and others use a dialect-specific code, as seen with Malay (\texttt{msa} vs \texttt{zlm}) and Odia (\texttt{ory} vs \texttt{ori}). These issues highlight the difficulties associated with crowd-sourced data collection efforts \cite{ardila2020common, conneau2023fleurs} and show the necessity of linguistic expertise during data development. As multilingual models become increasingly more common in both research and production \cite{whisper, mms, usm, seamless, chen2024towards}, proper linguistic meta-data is necessary to guarantee that models are configured properly.

After data cleaning, we segment the data into 3 splits: a baseline training set, a development set, and a test set. The baseline training set consists of 1 hour of data per language, while the development and test sets contain 10 minutes of data per language. The baseline training and development sets contain 139 languages, leading to a total of 139 hours of training data and 23 hours of development data. The test set contains an additional 10 languages from the ML-SUPERB Challenge hidden set \cite{mlsuperbchallenge}, totaling to 26 hours\footnote{Participants were made aware of which exact languages were included in the test set but not the training set.}.

\subsection{Accented and Dialectal Multilingual Data} \label{sec:dialectal_data}

To evaluate the performance of ASR models on different accents and dialects, we collect more evaluation data from new data sources, \textit{totaling to 93 accents and dialects}. We create 10-minute development and test subsets for each accent or dialect from each of these corpora in the same style as Section \ref{sec:data}. The development set contains data sourced from 9 accented or dialectal speech corpora~\cite{alharbi-etal-2024-sada, wang2021voxpopuli,demirsahin-etal-2020-open,wang24b_interspeech,zhao18b_interspeech,doganschönberger-2021-swissdial,vakirtzian24_interspeech,srivastava18_sltu,guevara-rukoz-etal-2020-crowdsourcing}. The hidden test set contains data sourced from the same corpora as the development set along with 4 additional corpora \cite{Abdallah-etal-2024-leveraging,casablanca,mdcc,hamalainen-etal-2021-finnish}. While we list these datasets here for transparency, participants were not made aware of which datasets were used during the challenge. We also did not reveal which accents or dialects were included in the challenge.

\vspace{-0.3cm}
\section{Challenge Task and Rules} \label{sec:benchmark}

\subsection{Updates from Previous Challenges and Benchmarks}
The goal of this challenge is to encourage the development of ASR systems that are robust to languages, accents, and dialects. Importantly, we avoid constraining participants to certain datasets or modeling approaches. This is distinct from the goals of previous ML-SUPERB benchmarks or challenges \cite{yang21c_interspeech}. The original benchmark constrained systems to a fixed training set, as it was designed to probe the multilingual capabilities of SSL models and not develop SOTA ASR systems. The ML-SUPERB 1.0 Challenge \cite{mlsuperbchallenge} was proposed to extend the benchmark to new languages, but enforced the same training constraints. Finally, the ML-SUPERB 2.0 benchmark\footnote{We emphasize that \cite{shi2024mlsuperb20benchmarkingmultilingual} is a \textit{benchmark} while this paper implements its findings into a \textit{challenge}.} \cite{shi2024mlsuperb20benchmarkingmultilingual} was designed to consider differences in architectures and training strategies when evaluating multilingual SSL models and proposed metrics to evaluate model robustness. However, it was also constrained to a fixed training set and did not consider robustness to accents or dialects. The differences are summarized in Table \ref{tab:versions}.

\setlength{\textfloatsep}{6pt}
\begin{table}[]
    \centering
    \caption{\textbf{Comparison of different ML-SUPERB versions}.}
    \label{tab:versions}
    \resizebox{\linewidth}{!}{
    \begin{tabular}{l|cccc}
    \toprule
    Version  &  Langs. & Models &  Fairness & Dialectal\\
    \midrule
    1.0 Benchmark & 143 & SSL & \xmark & \xmark \\
    1.0 Challenge & 154 & SSL & \xmark & \xmark\\
    2.0 Benchmark & 142 & ASR+SSL & \cmark & \xmark\\
    2.0 Challenge (ours) & 149 & Any & \cmark & \cmark\\
    \bottomrule
    \end{tabular}}
\end{table}

\vspace{-0.1cm}
\subsection{Task}

Contrary to previous SUPERB benchmarks \cite{superbsg, avsuperb, mlsuperb, feng2023superb, mlsuperbchallenge, shi2024mlsuperb20benchmarkingmultilingual, yang21c_interspeech}, the ML-SUPERB 2.0 Challenge only features a single evaluation track. Participants are tasked with developing SOTA multilingual ASR systems for 154 languages. During the evaluation phase of the challenge, the participants are given audio files and asked to both transcribe the speech (ASR) and predict the language being spoken (LID prediction). We opted not to include more complex tasks, such as Spoken Language Understanding, as it would lead to many low-resource languages being excluded.

\subsection{Submission} \label{sec:submission}
\vspace{-0.1cm}
We use DynaBench \cite{dynabench} as the challenge submission platform, as it was successfully used in previous competitions that required the test set to remain completely hidden, such as the FLoReS-101 Large-Scale MT challenge at WMT 2021 \cite{wenzek-etal-2021-findings, wmt-2021-findings}. Rather than submitting decoded inference results, participants instead upload their model submission to DynaBench. DynaBench interfaces with the uploaded model via a standardized, pre-defined API (Listing \ref{fig:schema}) and then performs inference and scoring on the evaluation server. Participants are only able to view the final evaluation scores, and do not have access to the input audio, ground truth text, nor their model outputs. This allows the evaluation to remain fully blind and robust, alleviating issues with community benchmark overfitting. A summary of the submission process is visualized in Figure \ref{fig:dynabench}.
\vspace{-0.1cm}

\subsection{Allowed Models and Methods} \label{sec:allowed_models}
\vspace{-0.1cm}
Unlike the ML-SUPERB 1.0 challenge that constrained participants to fixed data or training methods \cite{mlsuperbchallenge}, participants are allowed to use almost any resource in this benchmark. The only rule is that inference for the final evaluation must be performed on our server without internet connection (See Section \ref{sec:submission}). Furthermore, the submitted model must be able to perform inference within the server's memory limitations (around 46 GB of GPU VRAM). 

These flexible constraints allow participants to use the latest pre-trained models, such as self-supervised speech encoders \cite{chen2024towards, mms}, supervised ASR models \cite{whisper, owsm, peng2024owsmv31betterfaster, seamless}, or even Large Language Models (LLMs) \cite{touvron2023llama}, while preventing submissions that only use API-based models \cite{gpt3}. We note that participants are allowed to use API-based solutions to aid their own model development, such as for the purpose of distillation or pseudo-labelling.

\setlength{\textfloatsep}{6pt}
\begin{listing}[]
\begin{minted}{python}
def API(waveform):
    """
    Args:
        waveform (np.array): speech waveform
    Returns:
        pred_lid (str): ISO3 code of LID pred
        pred_asr (str): predicted transcript
    """
    
    # participants must fill in the below
    # lines to connect their model
    pred_lid = 
    pred_asr = 
    return pred_lid, pred_asr
\end{minted}
\caption{Pseudo-code for the API used to interface with the DynaBench, which participants customize for their model.}
\label{fig:schema}
\end{listing}

\vspace{-0.1cm}
\subsection{Training and Development Data}
\vspace{-0.1cm}
Similar to Section \ref{sec:allowed_models}, we do not have any explicit constraints on which datasets are allowed: participants are free to use an existing dataset for model development. They are also allowed to collect external data from new resources, as long as they mention how that data was obtained. To ease model development, we provide participants with the baseline training and development set based on the ML-SUPERB 2.0 public set \cite{shi2024mlsuperb20benchmarkingmultilingual}, as described in Section \ref{sec:multilingual_data}. We also provide participants with a secondary accented and dialectal development set (Section \ref{sec:dialectal_data}) to help them assess the robustness of their models to different language varieties. Participants are not required to use any of the provided data.
\vspace{-0.3cm}
\subsection{Evaluation Data and Metrics} \label{sec:metrics}
\vspace{-0.1cm}

\setlength{\textfloatsep}{6pt}
\begin{figure}
    \centering
    \includegraphics[width=1\linewidth]{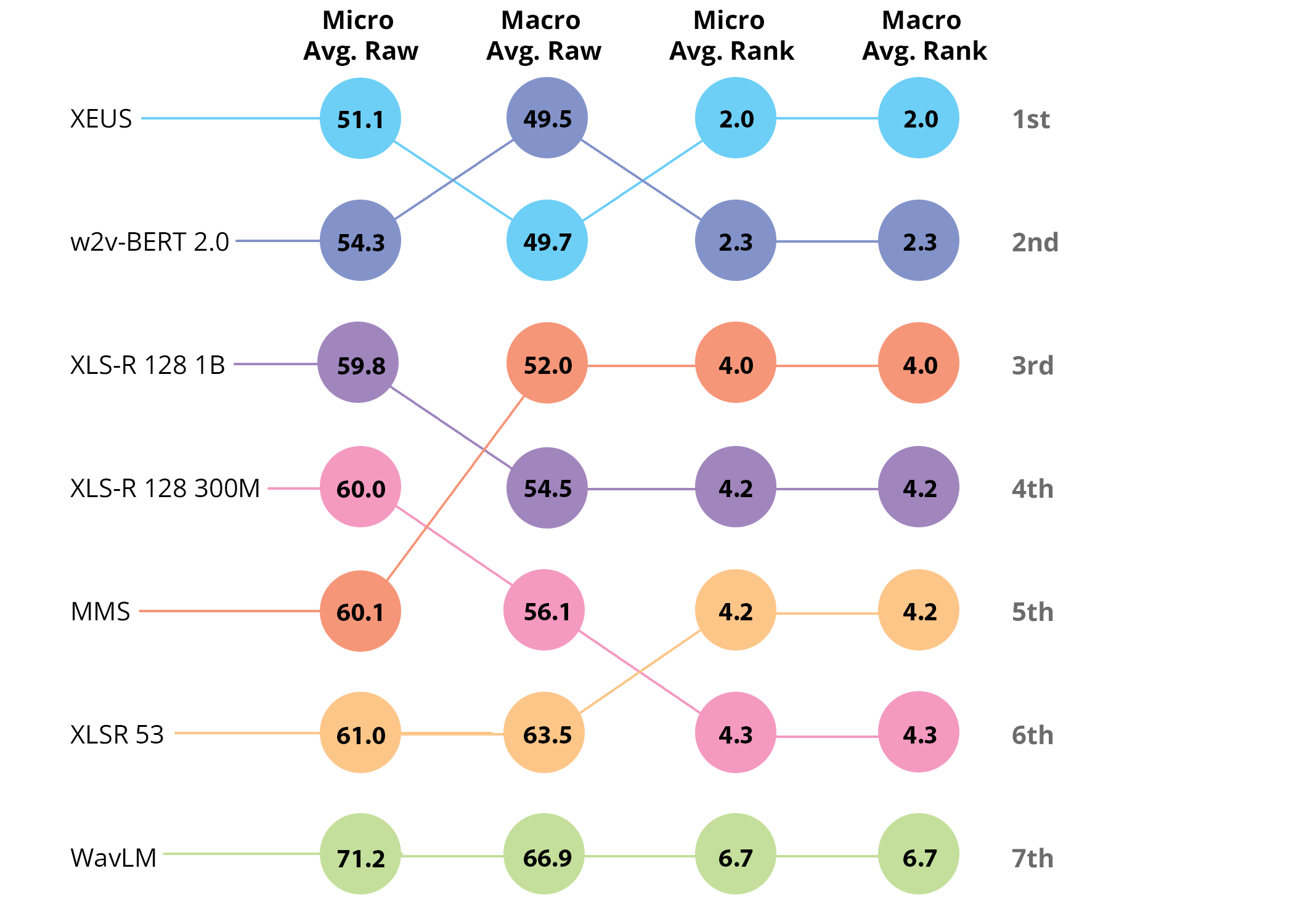}
    \vspace{-0.3cm}
    \caption{\textbf{Comparison of different aggregation methods used to compute the final rankings across all metrics.} Accuracies are inverted by subtracting from 100 when calculating the aggregate raw scores, such that a lower score is better.}

    \label{fig:avgs}
\end{figure}

The submissions of participants are evaluated on the accuracy of their speech recognition systems on different languages and language varieties. We separate our metrics into three categories: standard, language robustness, and dialectal robustness. Each of these categories are expanded upon in detail below.

\noindent \textbf{Standard: } This category mostly follows the original scoring system of ML-SUPERB's multilingual ASR+LID task with two main metrics: average language identification accuracy (ACC) and ASR character error rate (CER) across languages. The scores for this category are obtained from the inference results on the ML-SUPERB test sets discussed in Section \ref{sec:multilingual_data}.

\noindent \textbf{Language Robustness: } The language robustness metrics are designed to encourage consistency in performance across languages.\footnote{These metrics were first explored in the ML-SUPERB 2.0 benchmark \cite{shi2024mlsuperb20benchmarkingmultilingual}, but were not considered for model ranking.}  We use two metrics to measure this: average ASR CER of a system's worst performing 15 languages and the standard deviation (StD) of a system's ASR CER across all languages. These scores are also obtained from the inference results on the ML-SUPERB test sets discussed in Section \ref{sec:multilingual_data}. For both metrics, a lower score indicates a better result. 

\noindent \textbf{Dialectal Robustness: } The dialectal robustness evaluation metrics encourage systems that are robust to different varieties within languages. The metrics used for robustness mirror those of the standard evaluation category: language identification ACC and ASR CER on the accented and dialectal speech data. Participants are not required to identify the specific dialect spoken in an utterance. Since this evaluation requires accent/dialect labels, the scores are obtained from inference results on our new accented and dialectal test set discussed in Section~\ref{sec:dialectal_data}. 
\vspace{-0.3cm}
\subsection{Ranking}
\vspace{-0.1cm}
\begin{figure*}
    \centering
    \includegraphics[width=0.65\linewidth]{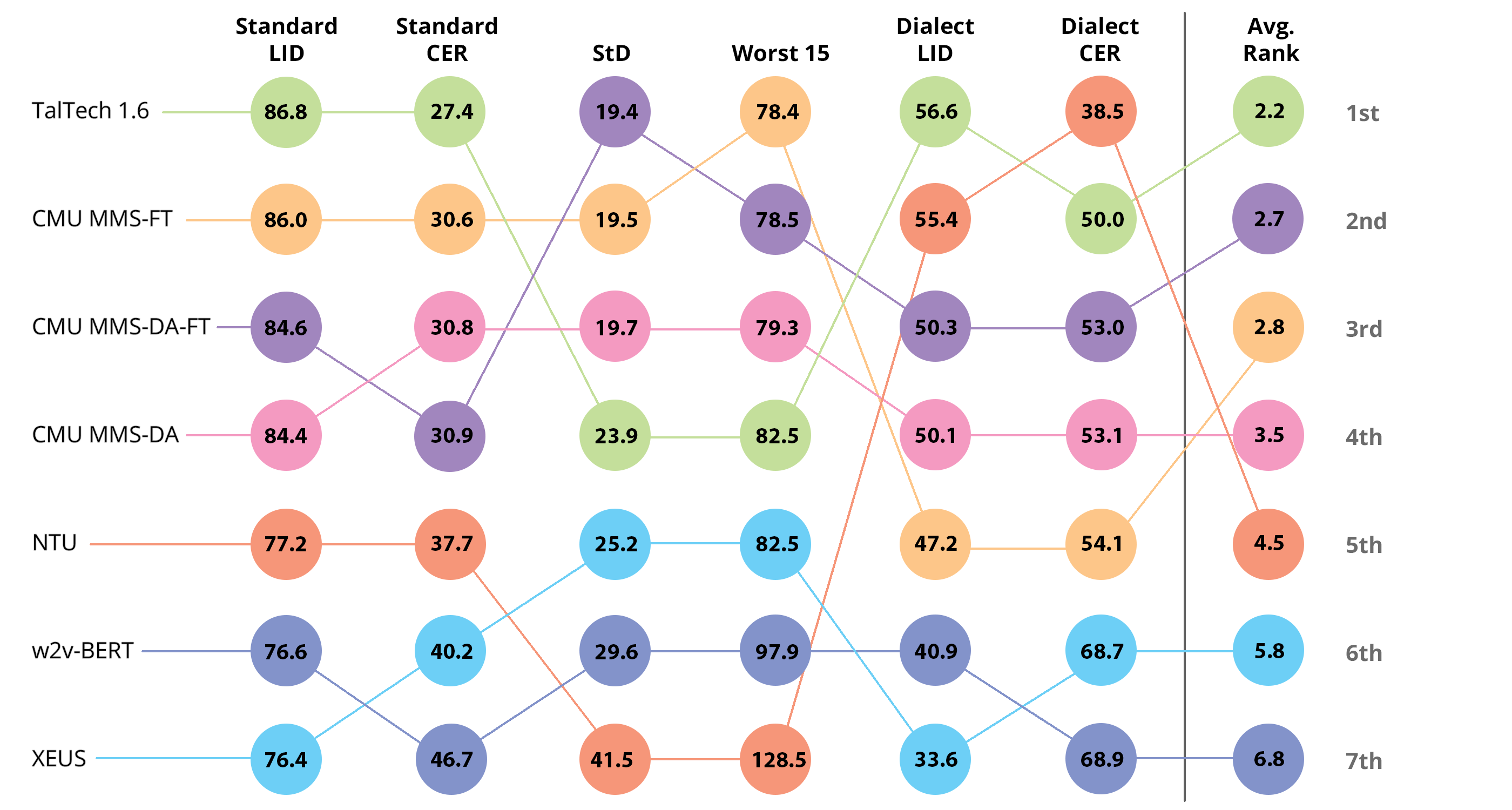}
    \caption{\textbf{Raw scores/rankings of each system submission and the best 2 SSL model baselines on our challenge evaluation metrics.}}
    \label{fig:overall_ranking}
    \vspace{-0.5cm}
\end{figure*}

The overall ranking on the challenge is calculated by taking the average rank of each model on each of the 6 metrics. The methodology can be summarized as follows:
\begin{enumerate}[left=12pt, before=\vspace{2pt}, after=\vspace{3pt}, itemsep=3pt]
    \item Calculate rankings for each model on each metric.
    \item Calculate the average ranking for each model across metrics.
    \item Rank models by average ranking.
\end{enumerate}

\noindent We use the average rank of a system rather than the average of the raw metric scores to avoid the ranking being skewed by differences in the metrics' dynamic ranges, which was effective in previous speech challenges \cite{chang2024interspeech, zhang2024neurips}. For example, the first two columns in Figure \ref{fig:avgs} show how rankings can shift significantly between a macro-average across the 3 categories (Standard, Language Robustness, Dialectal Robustness) and micro-average across the 6 individual metrics (Standard CER, Standard LID, Worst 15 CER, StD, Dialectal CER, Dialectal LID) when using raw scores, which is undesirable. On the other hand, there is no change in the overall ranking for the 7 SSL models between a macro-average and micro-average when using per-metric rank (Figure \ref{fig:avgs}, left 2 columns). We therefore use the micro-average across per-metric ranks as the aggregation method for simplicity and consistency, with tie-breakers decided by the micro-average of raw scores.

\vspace{-0.1cm}
\section{Benchmark and Submission Results} \label{sec:experiments}

\noindent\textbf{Self-Supervised Models: } We obtained baseline results with 7 systems based on SSL speech encoders: WavLM \cite{chen2022wavlm}, XLSR-53 \cite{conneau2020unsupervised}, XLS-R 128 300M \cite{babu2021xls}, XLS-R 128 1B \cite{babu2021xls}, MMS 1B \cite{mms}, w2v-BERT 2.0 \cite{seamless}, and XEUS \cite{chen2024towards}. Since all of these models are self-supervised, we develop ASR systems via fine-tuning on the ML-SUPERB 2.0 public set \cite{shi2024mlsuperb20benchmarkingmultilingual}. Each self-supervised speech encoder is frozen during fine-tuning, and a weighted sum of the layer-wise representations is used as input into a 2-layer Transformer encoder \cite{transformer} that is trained using the CTC loss \cite{ctc}.

\begin{table}[tb]
    \centering
    \caption{\textbf{Raw scores of SOTA supervised and SSL models on the ML-SUPERB and Dialectal test sets.}}
    \vspace{-0.3cm}
    \resizebox{\linewidth}{!}{
    \begin{tabular}{l|rr|rr|rr}
    \toprule
    & \multicolumn{2}{c}{Standard} & \multicolumn{2}{c}{Language} & \multicolumn{2}{c}{Dialectal} \\
    Model &  LID   & CER  & StD  & Worst  & LID & CER  \\
    \midrule
    WavLM & 55.9	& 72.0	& 38.3	& 122.1	& 27.2	& 79.0\\
    XLSR 53 & 53.8	& 51.9	& \textbf{19.4}	& 92.8	& 18.8	& 74.5\\
    XLS-R 300M & 72.6	& 53.0	& 31.4	& 104.3	& 30.5	& 74.5  \\
    XLS-R 1B   & 71.0	& 53.5	& 31.4	& 110.0	& 32.7	& 68.0  \\
    MMS 1B  & 74.6	& 57.1	& 33.5	& 118.8	& 40.4	& 65.9 \\
    w2v-BERT & \textbf{76.6}	& 46.7	& 29.6	& 97.9	& 40.9	& 68.9\\
    XEUS   & 76.4	& \textbf{40.2}	& 25.2	& \textbf{82.5}	& 33.6	& 68.7\\
    \midrule
    Whisper & 42.3 & 43.2 & 41.0 & 122.9 & \textbf{54.7} &\textbf{ 44.3} \\
    OWSM & 64.8 & 43.0 & 50.0 & 122.8 & 51.0 & 109.5\\
    \bottomrule
    \end{tabular}}
    \label{tab:scores}
\end{table}

\noindent\textbf{Supervised Models: } We also evaluate the capabilities of 2 multilingual pre-trained ASR foundation models: OWSM 4B \cite{chen2025owlsscalinglawsmultilingual} and Whisper Large v3 \cite{whisper}. These are run in a \textit{zero-shot} manner, as they are designed to be used out-of-the-box.
\vspace{-0.1cm}
\subsection{Results}
\vspace{-0.1cm}
\noindent \textbf{Supervised vs Self-Supervised: }
Table \ref{tab:scores} shows the scores of each supervised and SSL model. We find that SSL models generally achieve better scores due to the wide coverage of the fine-tuning dataset; supervised models yield inconsistent performances (high StD) since their original training set does not cover all languages. XEUS is the best overall model, while the English-only WavLM yields the worst results. Table~\ref{tab:unseen} breaks down the two supervised models' scores, separating languages seen in their training set from those that are unseen, showing the substantial impact of unseen languages on CER.

\begin{table}[tb]
    \centering
    \caption{\textbf{CERs of supervised models when considering if languages were seen during training.}}
    \label{tab:unseen}
    \vspace{-0.3cm}
    \begin{tabular}{l|rr|rr}
    \toprule
   & \multicolumn{2}{c}{Standard} & \multicolumn{2}{c}{Dialectal} \\
    Model &  Seen  & Unseen  & Seen & Unseen \\
    \midrule
    Whisper  & 40.7& 76.2 & \textbf{28.0} & \textbf{59.9} \\
    OWSM & \textbf{26.0} & \textbf{69.3} & 102.6 & 157.7\\
    \bottomrule
    \end{tabular}
\end{table}

\noindent \textbf{System Submissions: } The 5 systems from the 3 teams are compared to the 2 best SSL baselines in Figure~\ref{fig:overall_ranking}. All submitted systems achieved a higher overall ranking than the SSL baselines, showing the progress made during the challenge. No submission obtained the best results across all 6 metrics: each team had a system submission that ranked 1st in at least 1 metric. Comparing the best system per metric to XEUS, the best SSL baseline, we observed an absolute improvement of 12.4, 19.3, 5.8, 4.1, 23.0, and 30.2 on Standard LID, Standard CER, StD, Worst 15 CER, Dialect LID, and Dialect CER, respectively.

\vspace{-0.2cm}
\section{Conclusion} \label{sec:conclusion}
We propose the ML-SUPERB 2.0 Challenge, a novel speech processing challenge with the goal of inclusive ASR benchmarking for a large number (200+) of languages, accents, and dialects. The challenge introduces a novel multilingual test suite of accented and dialect speech and uses new metrics to test the robustness of ASR systems. The challenge received 5 system submissions from 3 teams, all of which outperformed baselines derived from SOTA SSL models. We hope future work can build on this challenge and its findings to make speech technologies more accessible for all.\\

\noindent\textbf{Acknowledgments} \ 
Antonios Anastasopoulos is supported by the National Science Foundation (NSF)under awards IIS-2125466 and IIS-2327143. Parts of this work used the PSC Bridges2 system at and Delta system at NCSA through allocations CIS210014 and IRI120008P from the ACCESS program, supported by NSF grants \#2138259,\#:2138286, \#:2138307, \#:2137603, and \#:2138296.

\section{References}
{
\setlength\bibitemsep{1pt}
\printbibliography

@article{conneau2020unsupervised,
  title={Unsupervised cross-lingual representation learning for speech recognition},
  author={Conneau, Alexis and Baevski, Alexei and Collobert, Ronan and Mohamed, Abdelrahman and Auli, Michael},
  journal={arXiv preprint arXiv:2006.13979},
  year={2020}
}

@article{babu2021xls,
  title={XLS-R: Self-supervised cross-lingual speech representation learning at scale},
  author={Babu, Arun and Wang, Changhan and Tjandra, Andros and Lakhotia, Kushal and Xu, Qiantong and Goyal, Naman and Singh, Kritika and von Platen, Patrick and Saraf, Yatharth and Pino, Juan and others},
  journal={arXiv preprint arXiv:2111.09296},
  year={2021}
}

@article{baevski2020wav2vec,
  title={wav2vec 2.0: A framework for self-supervised learning of speech representations},
  author={Baevski, Alexei and Zhou, Yuhao and Mohamed, Abdelrahman and Auli, Michael},
  journal={Proc. NeurIPS},
  volume={33},
  pages={12449--12460},
  year={2020}
}

@article{hsu2021hubert,
  title={Hubert: Self-supervised speech representation learning by masked prediction of hidden units},
  author={Hsu, Wei-Ning and Bolte, Benjamin and Tsai, Yao-Hung Hubert and Lakhotia, Kushal and Salakhutdinov, Ruslan and Mohamed, Abdelrahman},
  journal={TASLP},
  volume={29},
  pages={3451--3460},
  year={2021},
  publisher={IEEE}
}

@inproceedings{yang21c_interspeech,
  author={Shuwen Yang and Po-Han Chi and Yung-Sung Chuang and Cheng-I Jeff Lai and Kushal Lakhotia and Yist Y. Lin and Andy T. Liu and Jiatong Shi and Xuankai Chang and Guan-Ting Lin and Tzu-Hsien Huang and Wei-Cheng Tseng and Ko-tik Lee and Da-Rong Liu and Zili Huang and Shuyan Dong and Shang-Wen Li and Shinji Watanabe and Abdelrahman Mohamed and Hung-yi Lee},
  title={{SUPERB: Speech Processing Universal PERformance Benchmark}},
  year=2021,
  booktitle={Proc. Interspeech},
  doi={10.21437/Interspeech.2021-1775}
}

@inproceedings{whisper,
  title={Robust speech recognition via large-scale weak supervision},
  author={Radford, Alec and Kim, Jong Wook and Xu, Tao and Brockman, Greg and McLeavey, Christine and Sutskever, Ilya},
  booktitle={International conference on machine learning},
  pages={28492--28518},
  year={2023},
  organization={PMLR}
}

@inproceedings{superbsg,
  title={{SUPERB-SG: Enhanced Speech processing Universal PERformance Benchmark for Semantic and Generative Capabilities}},
  author={Tsai, Hsiang-Sheng and others},
  booktitle={Proceedings of the 60th Annual Meeting of the Association for Computational Linguistics (Volume 1: Long Papers)},
  year={2022}
}

@article{mms,
  title={Scaling speech technology to 1,000+ languages},
  author={Pratap, Vineel and others},
  journal={JMLR},
  volume={25},
  number={97},
  pages={1--52},
  year={2024}
}

@INPROCEEDINGS{mlsuperbchallenge,
  author={Shi, Jiatong and others},
  booktitle={2023 IEEE Automatic Speech Recognition and Understanding Workshop (ASRU)}, 
  title={{Findings of the 2023 ML-SUPERB Challenge: Pre-Training And Evaluation Over More Languages And Beyond}}, 
  year={2023},
  volume={},
  number={},
  doi={10.1109/ASRU57964.2023.10389628}}

@inproceedings{dynabench,
    title = "Dynabench: Rethinking Benchmarking in {NLP}",
    author = "Kiela, Douwe  and
      Bartolo, Max  and
      Nie, Yixin  and
      Kaushik, Divyansh  and
      Geiger, Atticus  and
      Wu, Zhengxuan  and
      Vidgen, Bertie  and
      Prasad, Grusha  and
      Singh, Amanpreet  and
      Ringshia, Pratik  and
      Ma, Zhiyi  and
      Thrush, Tristan  and
      Riedel, Sebastian  and
      Waseem, Zeerak  and
      Stenetorp, Pontus  and
      Jia, Robin  and
      Bansal, Mohit  and
      Potts, Christopher  and
      Williams, Adina",
    booktitle = "Proceedings of the 2021 Conference of the North American Chapter of the Association for Computational Linguistics: Human Language Technologies",
    month = jun,
    year = "2021",
    address = "Online",
    publisher = "Association for Computational Linguistics",
    url = "https://aclanthology.org/2021.naacl-main.324",
    doi = "10.18653/v1/2021.naacl-main.324",
}

@INPROCEEDINGS{avsuperb,
  author={Tseng, Yuan and Berry, Layne and Chen, Yi-Ting and Chiu, I-Hsiang and Lin, Hsuan-Hao and Liu, Max and Peng, Puyuan and Shih, Yi-Jen and Wang, Hung-Yu and Wu, Haibin and Huang, Po-Yao and Lai, Chun-Mao and Li, Shang-Wen and Harwath, David and Tsao, Yu and Mohamed, Abdelrahman and Feng, Chi-Luen and Lee, Hung-Yi},
  booktitle={ICASSP 2024 - 2024 IEEE International Conference on Acoustics, Speech and Signal Processing (ICASSP)}, 
  title={{AV-SUPERB: A Multi-Task Evaluation Benchmark for Audio-Visual Representation Models}}, 
  year={2024},
  volume={},
  number={},
  doi={10.1109/ICASSP48485.2024.10445941}}

@inproceedings{iwslt,
    title = "{Findings} {of} {the} {IWSLT} 2023 {Evaluation} {Campaign}",
    author = {Agarwal, Milind  and
      Agrawal, Sweta  and
      Anastasopoulos, Antonios  and
      Bentivogli, Luisa  and
      Bojar, Ond{\v{r}}ej  and
      Borg, Claudia  and
      Carpuat, Marine  and
      Cattoni, Roldano  and
      Cettolo, Mauro  and
      Chen, Mingda  and
      Chen, William  and
      Choukri, Khalid  and
      Chronopoulou, Alexandra  and
      Currey, Anna  and
      Declerck, Thierry  and
      Dong, Qianqian  and
      Duh, Kevin  and
      Est{\`e}ve, Yannick  and
      Federico, Marcello  and
      Gahbiche, Souhir  and
      Haddow, Barry  and
      Hsu, Benjamin  and
      Mon Htut, Phu  and
      Inaguma, Hirofumi  and
      Javorsk{\'y}, D{\'a}vid  and
      Judge, John  and
      Kano, Yasumasa  and
      Ko, Tom  and
      Kumar, Rishu  and
      Li, Pengwei  and
      Ma, Xutai  and
      Mathur, Prashant  and
      Matusov, Evgeny  and
      McNamee, Paul  and
      P. McCrae, John  and
      Murray, Kenton  and
      Nadejde, Maria  and
      Nakamura, Satoshi  and
      Negri, Matteo  and
      Nguyen, Ha  and
      Niehues, Jan  and
      Niu, Xing  and
      Kr. Ojha, Atul  and
      E. Ortega, John  and
      Pal, Proyag  and
      Pino, Juan  and
      van der Plas, Lonneke  and
      Pol{\'a}k, Peter  and
      Rippeth, Elijah  and
      Salesky, Elizabeth  and
      Shi, Jiatong  and
      Sperber, Matthias  and
      St{\"u}ker, Sebastian  and
      Sudoh, Katsuhito  and
      Tang, Yun  and
      Thompson, Brian  and
      Tran, Kevin  and
      Turchi, Marco  and
      Waibel, Alex  and
      Wang, Mingxuan  and
      Watanabe, Shinji  and
      Zevallos, Rodolfo},
    booktitle = "Proc. IWSLT",
    month = jul,
    year = "2023",
    address = "Toronto, Canada (in-person and online)",
    publisher = "Association for Computational Linguistics",
    url = "https://aclanthology.org/2023.iwslt-1.1",
    doi = "10.18653/v1/2023.iwslt-1.1",
}

@article{owsm,
  title={OWSM v3. 1: Better and faster open whisper-style speech models based on e-branchformer},
  author={Peng, Yifan and Tian, Jinchuan and Chen, William and Arora, Siddhant and Yan, Brian and Sudo, Yui and Shakeel, Muhammad and Choi, Kwanghee and Shi, Jiatong and Chang, Xuankai and others},
  journal={arXiv preprint arXiv:2401.16658},
  year={2024}
}

@article{seamless,
  title={SeamlessM4T-Massively Multilingual \& Multimodal Machine Translation},
  author={Barrault, Lo{\"\i}c and Chung, Yu-An and Meglioli, Mariano Cora and Dale, David and Dong, Ning and Duquenne, Paul-Ambroise and Elsahar, Hady and Gong, Hongyu and Heffernan, Kevin and Hoffman, John and others},
  journal={arXiv preprint arXiv:2308.11596},
  year={2023}
}

@article{usm,
  title={Google usm: Scaling automatic speech recognition beyond 100 languages},
  author={Zhang, Yu and Han, Wei and Qin, James and Wang, Yongqiang and Bapna, Ankur and Chen, Zhehuai and Chen, Nanxin and Li, Bo and Axelrod, Vera and Wang, Gary and others},
  journal={arXiv preprint arXiv:2303.01037},
  year={2023}
}

@inproceedings{mlsuperb,
  author={Jiatong Shi and Dan Berrebbi and William Chen and En-Pei Hu and Wei-Ping Huang and Ho-Lam Chung and Xuankai Chang and Shang-Wen Li and Abdelrahman Mohamed and Hung-yi Lee and Shinji Watanabe},
  title={{ML-SUPERB: Multilingual Speech Universal PERformance Benchmark}},
  year=2023,
  booktitle={Proc. INTERSPEECH},
  doi={10.21437/Interspeech.2023-1316},
  issn={2958-1796}
}

@inproceedings{ardila2020common,
  title={Common Voice: A Massively-Multilingual Speech Corpus},
  author={Ardila, Rosana and Branson, Megan and Davis, Kelly and Kohler, Michael and Meyer, Josh and Henretty, Michael and Morais, Reuben and Saunders, Lindsay and Tyers, Francis and Weber, Gregor},
  booktitle={Proc. LREC},
  pages={4218--4222},
  year={2020}
}

@article{accentarch,
  title={{Speech Accent Archive}},
  author={Weinberger, Steven},
  journal={Retrieved from http://accent.gmu.edu },
  year={2015}
}

@article{chen2024towards,
  title={Towards robust speech representation learning for thousands of languages},
  author={Chen, William and Zhang, Wangyou and Peng, Yifan and Li, Xinjian and Tian, Jinchuan and Shi, Jiatong and Chang, Xuankai and Maiti, Soumi and Livescu, Karen and Watanabe, Shinji},
  journal={arXiv preprint arXiv:2407.00837},
  year={2024}
}

@inproceedings{artie,
    title = "Artie Bias Corpus: An Open Dataset for Detecting Demographic Bias in Speech Applications",
    author = "Meyer, Josh  and
      Rauchenstein, Lindy  and
      Eisenberg, Joshua D.  and
      Howell, Nicholas",
    editor = "Calzolari, Nicoletta  and
      B{\'e}chet, Fr{\'e}d{\'e}ric  and
      Blache, Philippe  and
      Choukri, Khalid  and
      Cieri, Christopher  and
      Declerck, Thierry  and
      Goggi, Sara  and
      Isahara, Hitoshi  and
      Maegaard, Bente  and
      Mariani, Joseph  and
      Mazo, H{\'e}l{\`e}ne  and
      Moreno, Asuncion  and
      Odijk, Jan  and
      Piperidis, Stelios",
    booktitle = "Proceedings of the Twelfth Language Resources and Evaluation Conference",
    month = may,
    year = "2020",
    address = "Marseille, France",
    publisher = "European Language Resources Association",
    url = "https://aclanthology.org/2020.lrec-1.796",
    pages = "6462--6468",
    language = "English",
    ISBN = "979-10-95546-34-4",
}

@inproceedings{wang2021voxpopuli,
  title={VoxPopuli: A Large-Scale Multilingual Speech Corpus for Representation Learning, Semi-Supervised Learning and Interpretation},
  author={Wang, Changhan and Riviere, Morgane and Lee, Ann and Wu, Anne and Talnikar, Chaitanya and Haziza, Daniel and Williamson, Mary and Pino, Juan and Dupoux, Emmanuel},
  booktitle={Proceedings of the 59th Annual Meeting of the Association for Computational Linguistics and the 11th International Joint Conference on Natural Language Processing (Volume 1: Long Papers)},
  pages={993--1003},
  year={2021}
}

@inproceedings{conneau2023fleurs,
  title={Fleurs: Few-shot learning evaluation of universal representations of speech},
  author={Conneau, Alexis and Ma, Min and Khanuja, Simran and Zhang, Yu and Axelrod, Vera and Dalmia, Siddharth and Riesa, Jason and Rivera, Clara and Bapna, Ankur},
  booktitle={2022 IEEE Spoken Language Technology Workshop (SLT)},
  pages={798--805},
  year={2023},
  organization={IEEE}
}

@inproceedings{changAAVE,
  title={Self-supervised Speech Representations Still Struggle with African American Vernacular English},
  author={Chang, Kalvin and Chou, Yi-Hui and Shi, Jiatong and Chen, Hsuan-Ming and Holliday, Nicole and Scharenborg, Odette and Mortensen, David R},
  booktitle={INTERSPEECH},
  year={2024}
}

@article{feng2024towards,
  title={Towards inclusive automatic speech recognition},
  author={Feng, Siyuan and Halpern, Bence Mark and Kudina, Olya and Scharenborg, Odette},
  journal={Computer Speech \& Language},
  volume={84},
  pages={101567},
  year={2024},
  publisher={Elsevier}
}

@article{casablanca,
  title={Casablanca: Data and Models for Multidialectal Arabic Speech Recognition},
  author={Talafha, Bashar and Kadaoui, Karima and Magdy, Samar Mohamed and Habiboullah, Mariem and Chafei, Chafei Mohamed and El-Shangiti, Ahmed Oumar and Zayed, Hiba and Alhamouri, Rahaf and Assi, Rwaa and Alraeesi, Aisha and others},
  journal={arXiv preprint arXiv:2410.04527},
  year={2024}
}

@inproceedings{hamalainen-etal-2021-finnish,
    title = "{F}innish Dialect Identification: The Effect of Audio and Text",
    author = {H{\"a}m{\"a}l{\"a}inen, Mika  and
      Alnajjar, Khalid  and
      Partanen, Niko  and
      Rueter, Jack},
    editor = "Moens, Marie-Francine  and
      Huang, Xuanjing  and
      Specia, Lucia  and
      Yih, Scott Wen-tau",
    booktitle = "Proceedings of the 2021 Conference on Empirical Methods in Natural Language Processing",
    month = nov,
    year = "2021",
    address = "Online and Punta Cana, Dominican Republic",
    publisher = "Association for Computational Linguistics",
    url = "https://aclanthology.org/2021.emnlp-main.692",
    doi = "10.18653/v1/2021.emnlp-main.692",
    pages = "8777--8783"
}

@inproceedings{mdcc,
    title = "Automatic Speech Recognition Datasets in {C}antonese: A Survey and New Dataset",
    author = "Yu, Tiezheng  and
      Frieske, Rita  and
      Xu, Peng  and
      Cahyawijaya, Samuel  and
      Yiu, Cheuk Tung  and
      Lovenia, Holy  and
      Dai, Wenliang  and
      Barezi, Elham J.  and
      Chen, Qifeng  and
      Ma, Xiaojuan  and
      Shi, Bertram  and
      Fung, Pascale",
    editor = "Calzolari, Nicoletta  and
      B{\'e}chet, Fr{\'e}d{\'e}ric  and
      Blache, Philippe  and
      Choukri, Khalid  and
      Cieri, Christopher  and
      Declerck, Thierry  and
      Goggi, Sara  and
      Isahara, Hitoshi  and
      Maegaard, Bente  and
      Mariani, Joseph  and
      Mazo, H{\'e}l{\`e}ne  and
      Odijk, Jan  and
      Piperidis, Stelios",
    booktitle = "Proceedings of the Thirteenth Language Resources and Evaluation Conference",
    month = jun,
    year = "2022",
    address = "Marseille, France",
    publisher = "European Language Resources Association",
    url = "https://aclanthology.org/2022.lrec-1.696",
    pages = "6487--6494"
}

@inproceedings{feng2023superb,
  title={SUPERB@ SLT 2022: Challenge on Generalization and Efficiency of Self-Supervised Speech Representation Learning},
  author={Feng, Tzu-hsun and Dong, Annie and Yeh, Ching-Feng and Yang, Shu-wen and Lin, Tzu-Quan and Shi, Jiatong and Chang, Kai-Wei and Huang, Zili and Wu, Haibin and Chang, Xuankai and others},
  booktitle={2022 IEEE Spoken Language Technology Workshop (SLT)},
  pages={1096--1103},
  year={2023},
  organization={IEEE}
}

@article{chen2022wavlm,
  title={Wavlm: Large-scale self-supervised pre-training for full stack speech processing},
  author={Chen, Sanyuan and Wang, Chengyi and Chen, Zhengyang and Wu, Yu and Liu, Shujie and Chen, Zhuo and Li, Jinyu and Kanda, Naoyuki and Yoshioka, Takuya and Xiao, Xiong and others},
  journal={JSTSP},
  volume={16},
  number={6},
  pages={1505--1518},
  year={2022},
  publisher={IEEE}
}

@inproceedings{shi2024mlsuperb20benchmarkingmultilingual,
      title={{ML-SUPERB 2.0: Benchmarking Multilingual Speech Models Across Modeling Constraints, Languages, and Datasets}}, 
      author={Jiatong Shi and others},
      year={2024},
      booktitle={Proc. INTERSPEECH}
}

@inproceedings{peng2024owsmv31betterfaster,
      title={{OWSM v3.1: Better and Faster Open Whisper-Style Speech Models based on E-Branchformer}}, 
      author={Yifan Peng and others},
      year={2024},
      booktitle={Proc. INTERSPEECH}
}

@inproceedings{vakirtzian24_interspeech,
  title     = {Speech Recognition for Greek Dialects: A Challenging Benchmark},
  author    = {Socrates Vakirtzian and Chara Tsoukala and Stavros Bompolas and Katerina Mouzou and Vivian Stamou and Georgios Paraskevopoulos and Antonios Dimakis and Stella Markantonatou and Angela Ralli and Antonios Anastasopoulos},
  year      = {2024},
  booktitle = {Interspeech 2024},
  pages     = {3974--3978},
  doi       = {10.21437/Interspeech.2024-2443},
  issn      = {2958-1796},
}

@inproceedings{ctc,
  title={Connectionist temporal classification: labelling unsegmented sequence data with recurrent neural networks},
  author={Graves, Alex and Fern{\'a}ndez, Santiago and Gomez, Faustino and Schmidhuber, J{\"u}rgen},
  booktitle={ICML 2006},
  pages={369--376},
  year={2006}
}

@inproceedings{transformer,
  author       = {Ashish Vaswani and
                  Noam Shazeer and
                  Niki Parmar and
                  Jakob Uszkoreit and
                  Llion Jones and
                  Aidan N. Gomez and
                  Lukasz Kaiser and
                  Illia Polosukhin},
  title        = {Attention is All you Need},
  booktitle    = {NeurIPS 2017},
  year         = {2017},
}

@INPROCEEDINGS{alharbi-etal-2024-sada,
  author={Alharbi, Sadeen and Alowisheq, Areeb and Tüske, Zoltán and Darwish, Kareem and Alrajeh, Abdullah and Alrowithi, Abdulmajeed and Tamran, Aljawharah Bin and Ibrahim, Asma and Aloraini, Raghad and Alnajim, Raneem and Alkahtani, Ranya and Almuasaad, Renad and Alrasheed, Sara and Alsubaie, Shaykhah and Alonaizan, Yaser},
  booktitle={ICASSP 2024 - 2024 IEEE International Conference on Acoustics, Speech and Signal Processing (ICASSP)}, 
  title={SADA: Saudi Audio Dataset for Arabic}, 
  year={2024},
  volume={},
  number={},
  pages={10286-10290},
  doi={10.1109/ICASSP48485.2024.10446243}}

@inproceedings{demirsahin-etal-2020-open,
    title = "Open-source Multi-speaker Corpora of the {E}nglish Accents in the {B}ritish Isles",
    author = "Demirsahin, Isin  and
      Kjartansson, Oddur  and
      Gutkin, Alexander  and
      Rivera, Clara",
    editor = "Calzolari, Nicoletta  and
      B{\'e}chet, Fr{\'e}d{\'e}ric  and
      Blache, Philippe  and
      Choukri, Khalid  and
      Cieri, Christopher  and
      Declerck, Thierry  and
      Goggi, Sara  and
      Isahara, Hitoshi  and
      Maegaard, Bente  and
      Mariani, Joseph  and
      Mazo, H{\'e}l{\`e}ne  and
      Moreno, Asuncion  and
      Odijk, Jan  and
      Piperidis, Stelios",
    booktitle = "Proceedings of the Twelfth Language Resources and Evaluation Conference",
    month = may,
    year = "2020",
    address = "Marseille, France",
    publisher = "European Language Resources Association",
    url = "https://aclanthology.org/2020.lrec-1.804/",
    pages = "6532--6541",
    language = "eng",
    ISBN = "979-10-95546-34-4"
}

@inproceedings{wang24b_interspeech,
  title     = {GLOBE: A High-quality English Corpus with Global Accents for Zero-shot Speaker Adaptive Text-to-Speech},
  author    = {Wenbin Wang and Yang Song and Sanjay Jha},
  year      = {2024},
  booktitle = {Interspeech 2024},
  pages     = {1365--1369},
  doi       = {10.21437/Interspeech.2024-70},
  issn      = {2958-1796},
}

@inproceedings{zhao18b_interspeech,
  title     = {L2-ARCTIC: A Non-native English Speech Corpus},
  author    = {Guanlong Zhao and Sinem Sonsaat and Alif Silpachai and Ivana Lucic and Evgeny Chukharev-Hudilainen and John Levis and Ricardo Gutierrez-Osuna},
  year      = {2018},
  booktitle = {Interspeech 2018},
  pages     = {2783--2787},
  doi       = {10.21437/Interspeech.2018-1110},
  issn      = {2958-1796},
}

@misc{doganschönberger-2021-swissdial,
      title={SwissDial: Parallel Multidialectal Corpus of Spoken Swiss German}, 
      author={Pelin Dogan-Schönberger and Julian Mäder and Thomas Hofmann},
      year={2021},
      eprint={2103.11401},
      archivePrefix={arXiv},
      primaryClass={cs.CL},
      url={https://arxiv.org/abs/2103.11401}, 
}

@inproceedings{srivastava18_sltu,
  title     = {Interspeech 2018 Low Resource Automatic Speech Recognition Challenge for Indian Languages},
  author    = {Brij Mohan Lal Srivastava and Sunayana Sitaram and Rupesh {Kumar Mehta} and Krishna {Doss Mohan} and Pallavi Matani and Sandeepkumar Satpal and Kalika Bali and Radhakrishnan Srikanth and Niranjan Nayak},
  year      = {2018},
  booktitle = {6th Workshop on Spoken Language Technologies for Under-Resourced Languages (SLTU 2018)},
  pages     = {11--14},
  doi       = {10.21437/SLTU.2018-3},
}

@inproceedings{guevara-rukoz-etal-2020-crowdsourcing,
    title = "Crowdsourcing {L}atin {A}merican {S}panish for Low-Resource Text-to-Speech",
    author = "Guevara-Rukoz, Adriana  and
      Demirsahin, Isin  and
      He, Fei  and
      Chu, Shan-Hui Cathy  and
      Sarin, Supheakmungkol  and
      Pipatsrisawat, Knot  and
      Gutkin, Alexander  and
      Butryna, Alena  and
      Kjartansson, Oddur",
    editor = "Calzolari, Nicoletta  and
      B{\'e}chet, Fr{\'e}d{\'e}ric  and
      Blache, Philippe  and
      Choukri, Khalid  and
      Cieri, Christopher  and
      Declerck, Thierry  and
      Goggi, Sara  and
      Isahara, Hitoshi  and
      Maegaard, Bente  and
      Mariani, Joseph  and
      Mazo, H{\'e}l{\`e}ne  and
      Moreno, Asuncion  and
      Odijk, Jan  and
      Piperidis, Stelios",
    booktitle = "Proceedings of the Twelfth Language Resources and Evaluation Conference",
    month = may,
    year = "2020",
    address = "Marseille, France",
    publisher = "European Language Resources Association",
    url = "https://aclanthology.org/2020.lrec-1.801/",
    pages = "6504--6513",
    language = "eng",
    ISBN = "979-10-95546-34-4"
}

@INPROCEEDINGS{Abdallah-etal-2024-leveraging,
  author={Abdallah, Ahmed Amine Ben and Kabboudi, Ata and Kanoun, Amir and Zaiem, Salah},
  booktitle={ICASSP 2024 - 2024 IEEE International Conference on Acoustics, Speech and Signal Processing (ICASSP)}, 
  title={Leveraging Data Collection and Unsupervised Learning for Code-Switched Tunisian Arabic Automatic Speech Recognition}, 
  year={2024},
  volume={},
  number={},
  pages={12607-12611},
  doi={10.1109/ICASSP48485.2024.10445734}}

@inproceedings{wenzek-etal-2021-findings,
    title = "Findings of the {WMT} 2021 Shared Task on Large-Scale Multilingual Machine Translation",
    author = "Wenzek, Guillaume  and
      Chaudhary, Vishrav  and
      Fan, Angela  and
      Gomez, Sahir  and
      Goyal, Naman  and
      Jain, Somya  and
      Kiela, Douwe  and
      Thrush, Tristan  and
      Guzm{\'a}n, Francisco",
    editor = "Barrault, Loic  and
      Bojar, Ondrej  and
      Bougares, Fethi  and
      Chatterjee, Rajen  and
      Costa-jussa, Marta R.  and
      Federmann, Christian  and
      Fishel, Mark  and
      Fraser, Alexander  and
      Freitag, Markus  and
      Graham, Yvette  and
      Grundkiewicz, Roman  and
      Guzman, Paco  and
      Haddow, Barry  and
      Huck, Matthias  and
      Yepes, Antonio Jimeno  and
      Koehn, Philipp  and
      Kocmi, Tom  and
      Martins, Andre  and
      Morishita, Makoto  and
      Monz, Christof",
    booktitle = "Proc. WMT",
    month = nov,
    year = "2021",
    address = "Online",
    publisher = "Association for Computational Linguistics",
    url = "https://aclanthology.org/2021.wmt-1.2/",
    pages = "89--99"
}

@inproceedings{wmt-2021-findings,
    title = "Findings of the 2021 Conference on Machine Translation ({WMT}21)",
    author = "Akhbardeh, Farhad  and
      Arkhangorodsky, Arkady  and
      Biesialska, Magdalena  and
      Bojar, Ond{\v{r}}ej  and
      Chatterjee, Rajen  and
      Chaudhary, Vishrav  and
      Costa-jussa, Marta R.  and
      Espa{\~n}a-Bonet, Cristina  and
      Fan, Angela  and
      Federmann, Christian  and
      Freitag, Markus  and
      Graham, Yvette  and
      Grundkiewicz, Roman  and
      Haddow, Barry  and
      Harter, Leonie  and
      Heafield, Kenneth  and
      Homan, Christopher  and
      Huck, Matthias  and
      Amponsah-Kaakyire, Kwabena  and
      Kasai, Jungo  and
      Khashabi, Daniel  and
      Knight, Kevin  and
      Kocmi, Tom  and
      Koehn, Philipp  and
      Lourie, Nicholas  and
      Monz, Christof  and
      Morishita, Makoto  and
      Nagata, Masaaki  and
      Nagesh, Ajay  and
      Nakazawa, Toshiaki  and
      Negri, Matteo  and
      Pal, Santanu  and
      Tapo, Allahsera Auguste  and
      Turchi, Marco  and
      Vydrin, Valentin  and
      Zampieri, Marcos",
    editor = "Barrault, Loic  and
      Bojar, Ondrej  and
      Bougares, Fethi  and
      Chatterjee, Rajen  and
      Costa-jussa, Marta R.  and
      Federmann, Christian  and
      Fishel, Mark  and
      Fraser, Alexander  and
      Freitag, Markus  and
      Graham, Yvette  and
      Grundkiewicz, Roman  and
      Guzman, Paco  and
      Haddow, Barry  and
      Huck, Matthias  and
      Yepes, Antonio Jimeno  and
      Koehn, Philipp  and
      Kocmi, Tom  and
      Martins, Andre  and
      Morishita, Makoto  and
      Monz, Christof",
    booktitle = "Proc. WMT"
}

@article{touvron2023llama,
  title={{LLaMA}: Open and efficient foundation language models},
  author={Touvron, Hugo and Lavril, Thibaut and Izacard, Gautier and Martinet, Xavier and Lachaux, Marie-Anne and Lacroix, Timoth{\'e}e and Rozi{\`e}re, Baptiste and Goyal, Naman and Hambro, Eric and Azhar, Faisal and others},
  journal={arxiv:2302.13971},
  year={2023}
}

@inproceedings{gpt3,
 author = {Brown, Tom and Mann, Benjamin and Ryder, Nick and Subbiah, Melanie and Kaplan, Jared D and Dhariwal, Prafulla and Neelakantan, Arvind and Shyam, Pranav and Sastry, Girish and Askell, Amanda and Agarwal, Sandhini and Herbert-Voss, Ariel and Krueger, Gretchen and Henighan, Tom and Child, Rewon and Ramesh, Aditya and Ziegler, Daniel and Wu, Jeffrey and Winter, Clemens and Hesse, Chris and Chen, Mark and Sigler, Eric and Litwin, Mateusz and Gray, Scott and Chess, Benjamin and Clark, Jack and Berner, Christopher and McCandlish, Sam and Radford, Alec and Sutskever, Ilya and Amodei, Dario},
 booktitle = {Proc. NeurIPS},
 pages = {1877--1901},
 title = {Language Models are Few-Shot Learners},
 volume = {33},
 year = {2020}
}

@article{chang2024interspeech,
  title={The Interspeech 2024 Challenge on Speech Processing Using Discrete Units},
  author={Chang, Xuankai and Shi, Jiatong and Tian, Jinchuan and Wu, Yuning and Tang, Yuxun and Wu, Yihan and Watanabe, Shinji and Adi, Yossi and Chen, Xie and Jin, Qin},
  journal={arXiv preprint arXiv:2406.07725},
  year={2024}
}

@inproceedings{
zhang2024neurips,
title={Neur{IPS} 2024 Competition Proposal: {URGENT} Challenge},
author={Wangyou Zhang and Robin Scheibler and Kohei Saijo and Samuele Cornell and Chenda Li and Zhaoheng Ni and Anurag Kumar and Marvin Sach and Wei Wang and Shinji Watanabe and Tim Fingscheidt and Yanmin Qian},
booktitle={NeurIPS 2024 Competition Track},
year={2024},
url={https://openreview.net/forum?id=tVthtKNCGB}
}

@article{chen2025owlsscalinglawsmultilingual,
  title={{OWLS: Scaling Laws for Multilingual Speech Recognition and Translation Models}},
  author={William Chen and Jinchuan Tian and Yifan Peng and Brian Yan and Chao-Han Huck Yang and Shinji Watanabe},
  journal={arXiv preprint arXiv:2502.10373},
  year={2025}
}

@article{fairspeech,
  title={Towards measuring fairness in speech recognition: Fair-Speech dataset},
  author={Veliche, Irina-Elena and Huang, Zhuangqun and Kochaniyan, Vineeth Ayyat and Peng, Fuchun and Kalinli, Ozlem and Seltzer, Michael L},
  journal={arXiv preprint arXiv:2408.12734},
  year={2024}
}
}

\end{document}